\documentclass[runningheads]{llncs}
\usepackage{subcaption}

\usepackage[table,dvipsnames]{xcolor}
\usepackage[T1]{fontenc}
\usepackage{graphicx}
\usepackage{booktabs}
\usepackage[misc]{ifsym}

\usepackage{mwe}
\usepackage{hyperref}

\AtBeginDocument{%
  }

\usepackage{graphicx}
\usepackage{enumitem}
\usepackage{libertine}
\usepackage{booktabs}
\usepackage{tabularx}
\usepackage{multirow}
\newcolumntype{Y}{>{\raggedright\arraybackslash}X}
\usepackage[linesnumbered,ruled]{algorithm2e}
\usepackage{float}
\usepackage{subcaption}
\usepackage[most]{tcolorbox}
\usepackage{amsmath}
\usepackage{amssymb}
\usepackage{adjustbox}
\newtcolorbox{Box1}[2][]{
                lower separated=false,
                colback=white,
colframe=white!20!gray,fonttitle=\bfseries,
colbacktitle=white!10!gray,enhanced,
attach boxed title to top left={xshift=1cm,
        yshift=-2mm},
title=#2,#1}
\usepackage{graphicx, subcaption}
\usepackage{stfloats} 
\usepackage[misc]{ifsym}

\newcounter{xycomm}

\usepackage{booktabs} 
\newcounter{arkcomm}

\newcounter{xpcomm}

\newcounter{kxcomm}

\newcounter{todolist}

\begin{document}

\title{Order Acquisition Under Competitive Pressure: A Rapidly Adaptive Reinforcement Learning Approach for Ride-Hailing Subsidy Strategies}

\titlerunning{Order Acquisition Under Competitive Pressure}

\author{Fangzhou Shi \and
Xiaopeng Ke\and
Xinye Xiong\and
Kexin Meng\and 
Chang Men\and 
Zhengdan Zhu}
\tocauthor{Fangzhou Shi, Xiaopeng Ke, Xinye Xiong, Kexin Meng, Chang Men, Zhengdan Zhu} 
\toctitle{Order Acquisition Under Competitive Pressure: A Rapidly Adaptive Reinforcement Learning Approach for Ride-Hailing Subsidy Strategies}

\authorrunning{Shi et al.}

\institute{Didi Chuxing, Beijing, China \\ \email{\{arkshifangzhou,kexiaopeng,cintiaxiong,\\kexinmeng,menchang,zhuzhengdan\}@didiglobal.com}}

\maketitle              

\begin{abstract}
The proliferation of ride-hailing aggregator platforms presents significant growth opportunities for ride-service providers by increasing order volume and gross merchandise value (GMV). On most ride-hailing aggregator platforms, service providers that offer lower fares are ranked higher in listings and, consequently, are more likely to be selected by passengers. This competitive ranking mechanism creates a strong incentive for service providers to adopt coupon strategies that lower prices to secure a greater number of orders, as order volume directly influences their long-term viability and sustainability. Thus, designing an effective coupon strategy that can dynamically adapt to market fluctuations while optimizing order acquisition under budget constraints is a critical research challenge. However, existing studies in this area remain scarce.

To bridge this gap, we propose FCA-RL, a novel reinforcement learning-based subsidy strategy framework designed to rapidly adapt to competitors' pricing adjustments. Our approach integrates two key techniques: Fast Competition Adaptation (FCA), which enables swift responses to dynamic price changes, and Reinforced Lagrangian Adjustment (RLA), which ensures adherence to budget constraints while optimizing coupon decisions on new price landscape. Furthermore, we introduce RideGym, the first dedicated simulation environment tailored for ride-hailing aggregators, facilitating comprehensive evaluation and benchmarking of different pricing strategies without compromising real-world operational efficiency. Experimental results demonstrate that our proposed method consistently outperforms baseline approaches across diverse market conditions, highlighting its effectiveness in subsidy optimization for ride-hailing service providers.

\end{abstract}


\section{Introduction}

\begin{figure}[h]
    \centering
    \subfloat[\textbf{The interaction process between a passenger, a ride-hailing aggregator, and ride service providers}.]{
        \begin{minipage}{0.56\textwidth}
            \centering
            \includegraphics[width=\linewidth]{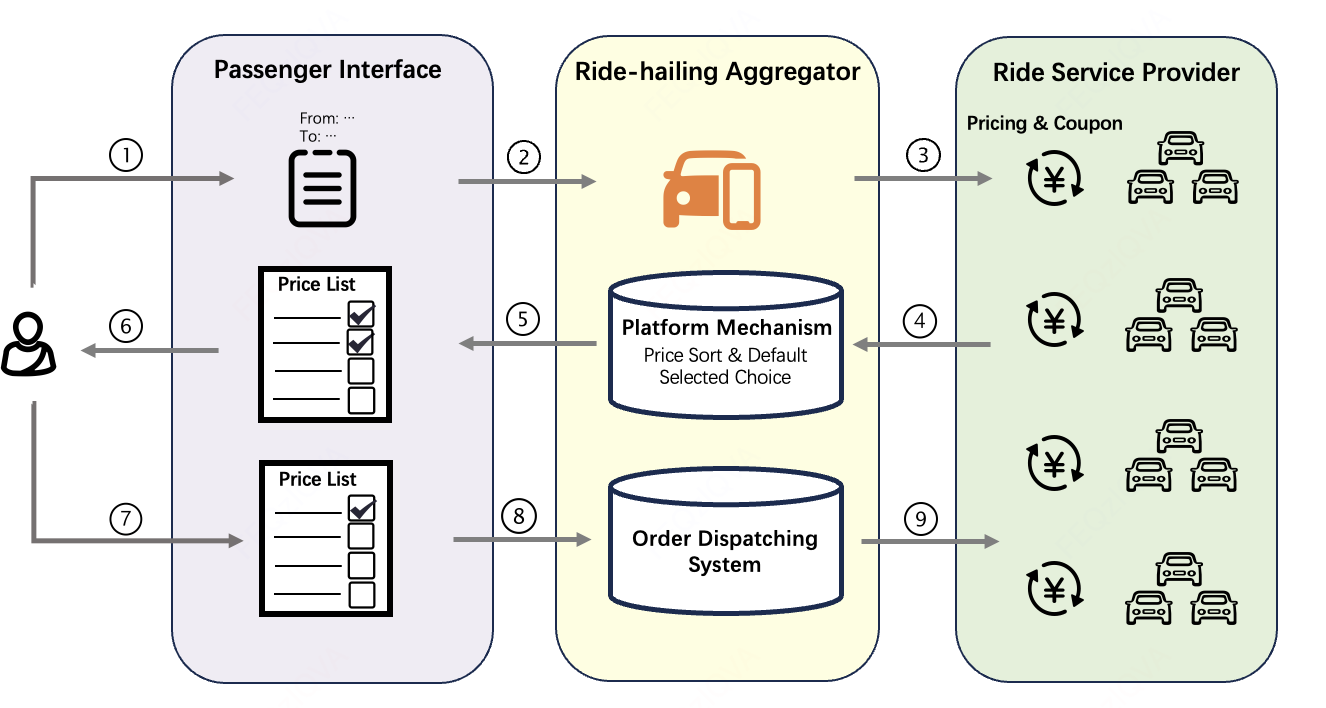}
            \caption*{
            \textcircled{\small{1}} - \textcircled{\small{3}} The passenger inputs the origin and destination, prompting the aggregator to request price quotes from all service providers.\\
            \textcircled{\small{4}} - \textcircled{\small{6}} Service providers submit their quoted prices after applying coupons. The aggregator ranks them and auto-selects certain options based on its mechanism before presenting the price list to the passenger.\\
            \textcircled{\small{7}} - \textcircled{\small{9}} The passenger send an order with the chosen options, and the aggregator dispatches it to the most suitable driver from the selected service providers.
            }
            \label{fig:all-pipeline}
        \end{minipage}
    }
    \hfill
    \subfloat[\textbf{A demonstration of the passenger's interface form on the ride-hailing aggregator}.]{
        \begin{minipage}{0.35\textwidth}
            \centering
            \adjustbox{max height=9.2cm}{
                \includegraphics[width=\linewidth]{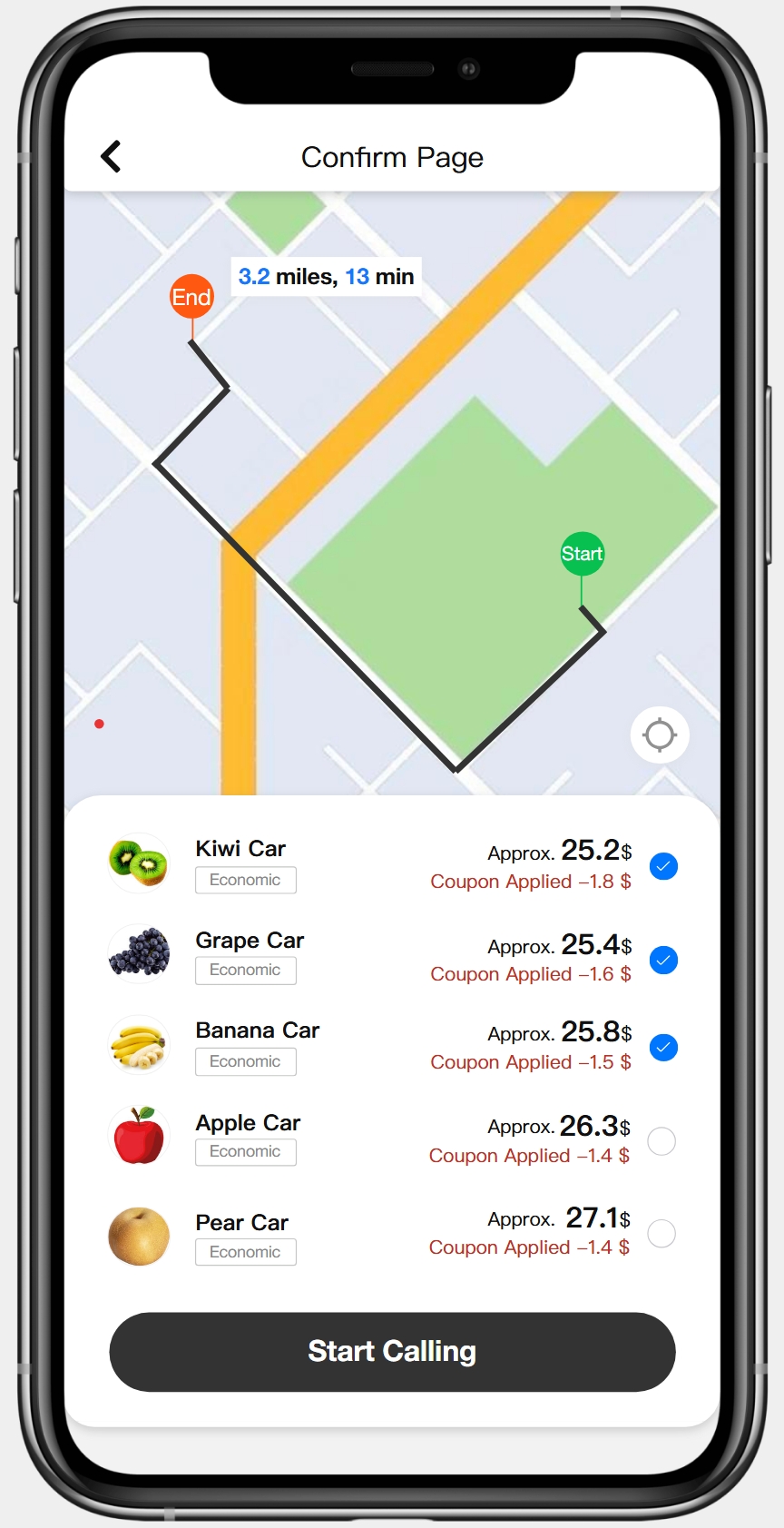}
            }
            \label{fig:form_demo}
        \end{minipage}
    }
    \caption{\textbf{The ride-hailing aggregation process and passenger interface.}}
    \label{fig:combined_fig}
\end{figure}

The rise of ride-hailing aggregators~\cite{bao2023mathematical}\textemdash platforms integrating various third-party ride-hailing services\textemdash has grown alongside the ride-hailing industry~\cite{wang2019ridesourcing}. These platforms enhance demand realization and social welfare by increasing market thickness and reducing matching frictions, allowing smaller ride-service providers to boost order volume and gross merchandise value (GMV). As depicted in the Figure~\ref{fig:all-pipeline}, key stakeholders include the \textit{Passenger}, \textit{Ride-Hailing Aggregators (RHA)}, and \textit{Ride-Service Providers (RSP)}. The prices quoted by each RSP are sorted and displayed on the passenger's interface (demonstrated in Figure~\ref{fig:form_demo}). To improve user experience, RHAs typically automatically select the top-K lowest-priced options. This mechanism strongly motivates
RSPs to lower their price to fall within this range, as the majority of passengers tend to maintain this default range when sending orders due to inherent inertia.

Once an order is sent, it will be dispatched to one of the most suitable drivers of the RSPs selected by the passenger. This means that, for a RSP,  the fewer ride-service options a passenger selects in a request, the less competition for order completion it will face. Therefore, competing on price (i.e., the travel fare) with other RSPs essentially prompts passengers to select and send orders, thereby increasing the likelihood of order completion. Unlike other studies that discuss long-term base price pricing strategies~\cite{wang2016pricing,zhou2021long}, or dynamic price surging~\cite{chen2016dynamic} under extreme supply \& demand conditions, this paper focuses on RSPs that use instant coupons to granularly adjust prices in response to price competition in RHA, a topic that has been underexplored in previous research.

The limited budget of each RSP necessitates the development of efficient coupon allocation strategies for competitive effectiveness, which presents a significant research challenge. Traditional methods borrow ideas from user marketing~\cite{zou2020heterogeneouscausallearningeffectiveness}, using uplift modeling~\cite{gutierrez2017causal} techniques to model the individual treatment effect in the passenger's willingness to send an order, after which the duality method with Lagrangian multiplier is applied to solve the maximization problem of order volume (or GMV) under budget constraints. However, these methods usually assume a stable distribution of features, treating the data as Independent and Identically Distributed (IID), which does not hold in the real world, especially when other RSPs perform aperiodic price adjustments.

The field of Real-Time Bidding (RTB)~\cite{zhang2014optimal} faces a similar challenge of dynamic price competition. In RTB, merchants bid for ad positions in personalized recommendations, akin to how RSPs use coupons to improve rankings in passenger requests. Dynamic price competition occurs in both areas, with some studies in RTB predicting market prices~\cite{ren2019deep} from historical data or using reinforcement learning for real-time bid adjustments~\cite{cai2017real,he2021unified,wang2022roi}. However, applying RTB methods to RHA is complicated by two main differences. First, RSP coupons directly affect the rank and the final travel fare, whereas RTB bids are not related to the final sales price. Second, order completion in RHA depends on the number of selected RSPs, real-time supply status, and the RHA dispatch algorithm, which introduces significant uncertainty. In contrast, the conversion rate in RTB depends primarily on the item price and user interest. These differences complicate the design of coupon strategies in the RHA environment.

We propose FCA-RL, a framework that uses Bayesian posterior updates and reinforcement learning to dynamically mitigate the adverse effects of competitors' aperiodic price adjustments on coupon efficiency and budget control. From the perspective of RSPs, we formulate this budget-constrained order maximization problem as a Markov Decision Process (MDP). To solve this MDP under aperiodic price adjustments by competitors, we introduce a \textbf{Fast Competition Adaptation (FCA)} module that quickly tracks the evolving price landscape, and a \textbf{Reinforced Lagrangian Adjustment (RLA)} module that optimizes coupon allocation according to the updated market conditions. The integration of these two modules ensures sustained efficiency and accurate budget execution in a dynamic RHA environment.

Evaluating coupon strategies in real time is risky due to potential declines in GMV and order volume. To overcome this, we developed \textbf{RideGym}, an offline simulation system that emulates the ride-hailing process and allows for flexible configurations, including supply factors and competition levels. In addition, this system serves as an interactive environment for training our reinforcement learning agent. Our evaluations show that our algorithm significantly outperforms baseline models by reducing budget control errors and improving the Finish Return on Investment.


Our main contributions are threefold:
 \textbf{(1) The first work focuses on the coupon strategies of RSPs in RHA.} To the best of our knowledge, we are the first study to explore using coupons for instant price competition in RHAs from the perspective of a RSP. 
\textbf{(2) A novel dynamic RL subsidy framework.} We propose FCA-RL, a novel reinforcement learning-based coupon strategy framework configured with a module for rapid market response in RHA, which enables precise budget control in dynamic markets without compromising coupon efficacy. 
\textbf{(3) A Full-Chain RHA Simulator.}We have developed a simulation system, \textbf{RideGym}, that models the complete lifecycle of an order within the RHA, facilitating comprehensive analysis of coupon strategies under various algorithmic and environmental variations.

\section{Related Works}
\label{sec:related-works}

\paragraph{Coupon Strategy in Industry.}Coupon strategies are widely employed in e-commerce to enhance customer purchasing willingness while adhering to budget constraints. Zou et al.~\cite{zou2020heterogeneouscausallearningeffectiveness} formulate this problem as a constrained optimization task based on uplift estimation and solve it using a duality method with Lagrangian relaxation, under the assumption of a stable distribution. Dai et al.~\cite{dai2024data} introduced a PID controller to dynamically adjust the Lagrangian multiplier for real-time traffic control but did not account for the optimality. Xiao et al.~\cite{xiao2023modelbasedconstrainedmdpbudget} define a model-based constrained Markov decision process (CMDP) and propose a joint optimization of both $\lambda$ and the policy. Zhang et al.~\cite{zhang2021bcorle} utilize offline reinforcement learning to allocate personalized discounts under a predefined budget, leveraging offline dataset augmentation with multiple $\lambda$ choices; however, their offline approach is unsuitable for scenarios where market conditions are unpredictable and evolve irregularly. Chen et al.~\cite{chen2024optimal} design an optimal instant discount strategy for multiple products on ride-hailing aggregators, but their focus on aggregator profitability diverges from our approach.

\paragraph{Online Advertising Bidding.} The optimal coupon strategy of RSP in RHA is similar to advertisers bidding in RTB, where merchants compete for prime placements. Zhang et al.~\cite{zhang2014optimal} model the winning function and advertisers' private values under specific formulations, deriving an optimal bidding strategy. Recently, reinforcement learning has gained traction in bidding. Cai et al.~\cite{cai2017real} model request distribution transitions in advertising as an MDP, using dynamic programming for optimal bidding. He et al.~\cite{he2021unified} propose a general bidding framework using the dual Lagrangian method and reinforcement learning for budget control under traffic fluctuations but remains ineffective for market price dynamics. Wang et al. \cite{wang2022roi,Wang_2023} model the problem as a Partial Observable Markov Decision Process (POMDP), leveraging Transformers and variational inference to capture market price uncertainties, achieving better adaptability.

\section{Problem Formalization}
\label{sec:problem-formulation}
\subsection{Original Problem}
\label{subsec: ori_problem}
\paragraph{Notation.} 
Suppose \( M \) RSPs participate in the RHA. From the perspective of a single RSP, we define a budget rate \( B \in [0,1] \) and a coupon set \( \mathbf{d} = \{d_0, \dots, d_{H-1}\} \subseteq [0,1] \), where a 20\% discount corresponds to \( d_j = 0.2 \). Over the period \([t_{start}, t_{end}]\), we collect a dataset \( \mathcal{D}_{t_{start}}^{t_{end}} = \{(\mathbf{x}_i, g_i, d_{i,j}, y^{I,j}_i, y^{C,j}_i)\}_{i=0}^{N-1} \), which includes contextual features \( \mathbf{x}_i \) (e.g., supply-demand conditions and order distance), base prices \( g_i \in \mathbb{R}_{+}^1 \) before coupon application, the applied coupon \( d_{i,j} \), and two binary variables: \( y^{I,j}_i \in \{0, 1\} \), indicating whether the RHA automatically selects the RSP in the passenger's form (Figure~\ref{fig:all-pipeline}-\textcircled{\small{5}}) with coupon \( d_{i,j} \), and \( y^{C,j}_i \in \{0, 1\} \), indicating the order completion status after applying \( d_{i,j} \). Notably, \( y^{I,j}_i \) and \( y^{C,j}_i \) are sampled from the corresponding potential outcome distributions \( P^{I,(j)}_{\textbf{x}_i} \) and \( P^{C,(j)}_{\textbf{x}_i} \) for coupon \( d_{i,j} \). Additionally, we introduce a new dataset \( \mathcal{D}_{t_{end}+1}^{t_{end}+L} \), where the potential outcome distributions \( P^{I,(j)}_{\textbf{x}} \) and \( P^{C,(j)}_{\textbf{x}} \) for each coupon \( d_j \) are influenced by competitors' sporadic price adjustments, leading to distributions that differ from those in \( \mathcal{D}_{t_{start}}^{t_{end}} \).

\paragraph{Original Formulation.} 
Given \( \mathcal{D}_{t_{start}}^{t_{end}} \), let \( z_{ij} \in [0,1] \) represent the counterfactual estimation of \( y_i^{C,(j)} \), which can be derived from a model trained on this dataset by uplift modeling techniques~\cite{gutierrez2017causal}, indicating the order completion probability for opportunity \( i \) when applying coupon \( d_j \) over \([t_{start}, t_{end}]\). We define \( v_{ij} \in \{0,1\} \) as the decision variable indicating whether coupon \( d_j \) is applied to opportunity \( i \), ensuring exactly one coupon per request: \( \mathbf{v}_i = (v_{i0}, \dots, v_{i(H-1)})^\intercal \in \{0,1\}^H \). Our objective is to optimize \( \{\mathbf{v}_i\}_{i=0}^{N-1} \) to maximize order completions while adhering to a predetermined budget rate, ensuring that the total coupon expenditure does not exceed the product of total earned GMV and the budget rate. Additionally, we aim to develop an adjustment algorithm to enable rapid adaptation to new data distributions in \( \mathcal{D}_{t_{end}+1}^{t_{end}+L} \). We begin by formulating a constrained optimization problem on the original dataset \(  \mathcal{D}_{t_{start}}^{t_{end}} \)

\begin{equation}
\begin{aligned}
   \mathop{min}\limits_{\mathbf{v}} & -\sum_{i=0}^{N-1} \sum_{j=0}^{H-1} z_{ij}v_{ij} & \\
   s.t. & \sum_{i=0}^{N-1} \sum_{j=0}^{H-1} g_i z_{ij} v_{ij} d_j  \leq \sum_{i=0}^{N-1} \sum_{j=0}^{H-1} g_i z_{ij}v_{ij} \cdot B \\
   & v_{ij} \in \{0,1\}, \quad \sum_{j=0}^{H-1} v_{ij}= 1 \\
   & \forall i \in \{0, \cdots, N-1\}, \quad \forall j \in \{0, \cdots, H-1\} \\
\end{aligned}
\label{eq:primaryeq}   
\end{equation}


\subsection{Optimal Decision Function}
The integrality constraint on $v$ is removed by applying linear relaxation. Other constraints are relaxed using the Lagrange multiplier, and the problem is then reformulated in its dual form (see Appendix C.1 for details):
\begin{equation}
\mathop{max}_{\lambda \geq 0} \: \mathop{min}_{\mathbf{v}} \ L(\mathbf{v}, \lambda) = \sum_{i=0}^{N-1} \sum_{j=0}^{H-1} z_{ij}(\lambda g_i d_j - \lambda g_i B -1 )v_{ij}
\label{eq:relaxedLagrangian}
\end{equation}
This Lagrangian dual transformation preserves the optimal value of the original problem through the strong duality (see Appendix C.2 for proof). From Eq.~\ref{eq:relaxedLagrangian}, the optimal discount $j^*(\lambda)$ for each opportunity $i$, given a fixed $\lambda$, is:
\begin{equation}
j^*(\lambda) = \mathop{argmin}\limits_j \ z_{ij}(\lambda g_i d_j - \lambda g_i B -1)
\label{eq:relaxedLagrangian-adjj}
\end{equation}
where $v_{ij} = 1$ for $j = j^*$ and $v_{ij} = 0$ otherwise (see Appendix C.3 for details). Given this formulation, the maximum of Eq.~\ref{eq:relaxedLagrangian} is obtained by tuning $\lambda$ over a sum of piecewise linear convex functions. To handle non-differentiable points, we apply a ternary search to iteratively determine $\lambda^*$. 

However, competitors' aperiodic price adjustments lead to continuous changes in the distributions of \( \mathbf{z} \), making the optimized \( \lambda^* \) from \( \mathcal{D}_{t_{start}}^{t_{end}} \) non-generalizable. To address this, we decompose the order completion probability \( z_{ij} \) (Eq.~\ref{eq:toatlprobabilityds}) into \( f^{(in)}_{ij} \) (completion probability when auto-selected) and \( f^{(out)}_{ij} \) (when not auto-selected). We explicitly model the RSP's In-Range Rate (\textit{IRR}, Def.~\ref{def: irr}) as \( w_{ij} \), representing the probability of entering the auto-selection range for each coupon, since the \textit{IRR} is the most sensitive part affected by the competitors' sporadic price changes, unlike \( f^{(in)}_{ij} \) and \( f^{(out)}_{ij} \). Thus, tracking and calibrating the \textit{IRR} distribution in real-time and applying adaptive optimization are critical to maintaining efficiency in \( \mathcal{D}_{t_{end}+1}^{t_{end}+L} \). We model this as a Markov Decision Process (MDP)~\cite{szepesvari2022algorithms}, allowing reinforcement learning to adjust \( \lambda \) at each time step in response to the real-time \textit{IRR} changes.
\begin{equation}
    z_{ij} = w_{ij}f^{(in)}_{ij} + (1-w_{ij})f^{(out)}_{ij}
\label{eq:toatlprobabilityds}
\end{equation}

\begin{definition}
 (In-Range Rate, IRR). \textit{For a ride-hailing opportunity \( i \), the IRR represents the probability that an RSP’s quoted price is auto-selected by the RHA to maximize user satisfaction. The number of In-Range RSPs depends on the opportunity's context \( \mathbf{x}_i \) and follows a top-\( K(\mathbf{x}_i) \) lowest-price rule, where \( K: \mathcal{X} \to \mathbb{N}^+ \) maps features to the number of auto-selections.}
 \label{def: irr}
\end{definition}

\section{Method}
\label{sec:method}
\begin{figure}[t]
    \centering
    \includegraphics[width=\linewidth]{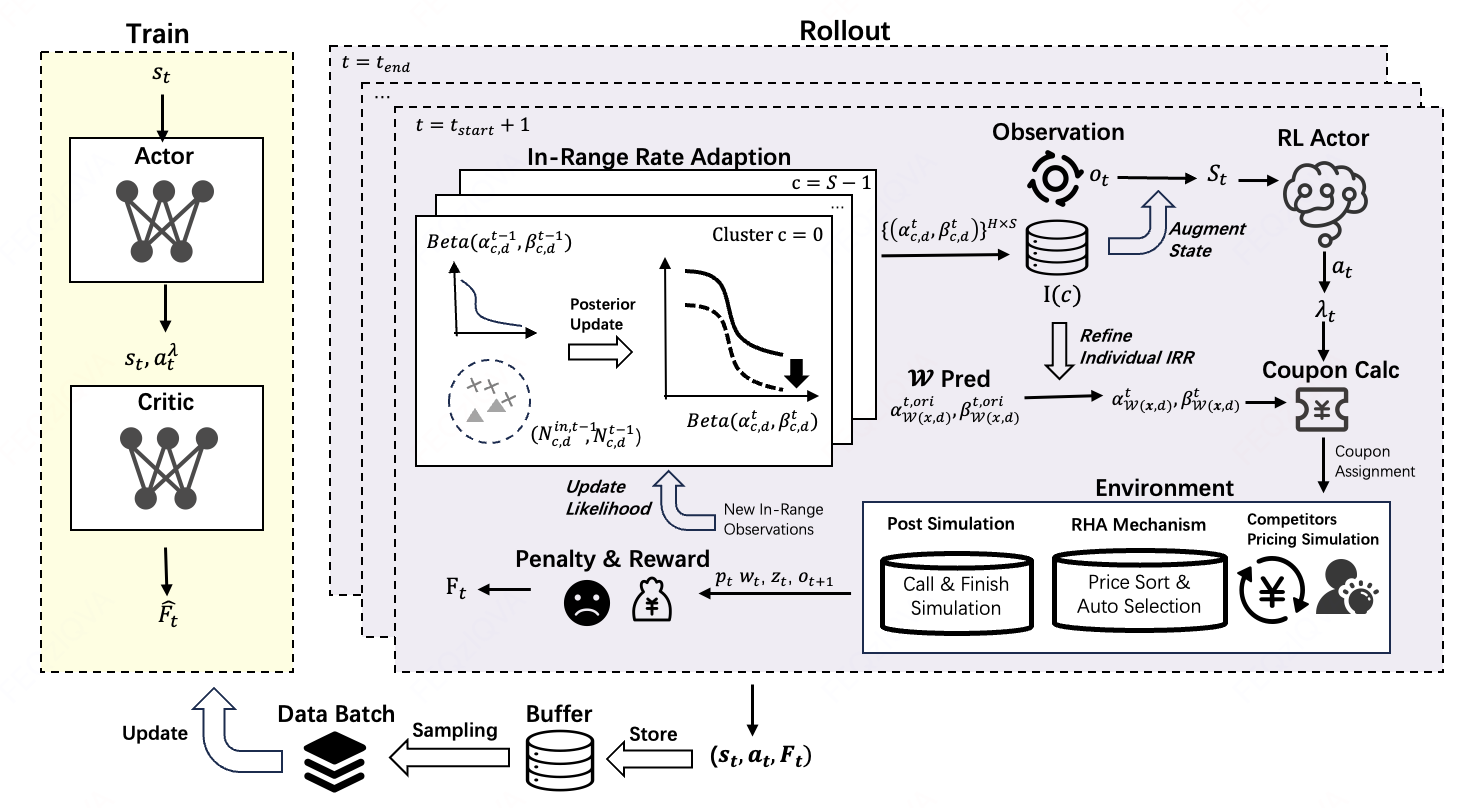}
    \caption{An overview of our FCA-RL framework which detailing the integration of FCA and RLA at each rollout step. After several rollouts, a batch of tuples $(s_t, a_t, F_t)$ is sampled for reinforcement learning training.}
    \label{fig:framework}
\end{figure}

In this section, we introduce FCA-RL, a reinforcement learning-based coupon strategy framework tailored for competitive RHA environments. We define a MDP as outlined in Section~\ref{subsec:mdp} and propose to solve it through our proposed \textbf{Reinforced Lagrangian Adjustment (RLA)}, with an online module, \textbf{Fast Competition Adaptation (FCA)}, integrated to monitor and adapt to the evolving IRR landscape influenced by competitors' sporadic price adjustments. An overview of our framework is demonstrated in Figure~\ref{fig:framework}.

\subsection{Reinforced Lagrangian Adjustment (RLA)}
\subsubsection{Markov Decision Process}
\label{subsec:mdp}
We model the $\lambda$ adaptation process as a finite-horizon MDP $(\mathcal{S}, \mathcal{A}, \mathcal{T}, \mathcal{R}, \mathcal{P}, \mathcal{F}, \mathcal{\gamma})$ with the state space $\mathcal{S}$, the action set $\mathcal{A}$, the transition function $\mathcal{T}$, the reward function $\mathcal{R}$, the penalty function $\mathcal{P}$, the full-achievement function $\mathcal{F}$, and the discount factor $\mathcal{\gamma}$. In our study, we equate the long-term impact by setting $\mathcal{\gamma} = 1$.

\paragraph{State.} Each state $s \in \mathcal{S}$ at time step $t$ encapsulates: a normalized time slot index, the previous Lagrangian multiplier \( \lambda_{t-1} \), the coupon campaign status up to $t$ (including the executed budget rate, gap to target, gained GMV, and Finish ROI), and statistics of the latest \textit{IRR} distribution. This state representation supports effective decision-making based on \textit{IRR} changes.

\paragraph{Action.} 
At each time step $t$, the action $\mathbf{a}_t \in \mathcal{A}$ adjusts $\lambda_{t-1}$ via Eq.~\ref{eq:act_aply}, with $\eta$ controlling the extent of the adjustment. The updated $\lambda_t$ is then used in Eq.~\ref{eq:relaxedLagrangian-adjj} for coupon assignment. $lb$ and $up$ are set to be the lower bound and the upper bound of $\lambda$.
\begin{equation}
\begin{aligned}
  \lambda_t = \min(\max(\lambda_{t-1}\cdot\eta\cdot a_t, lb), ub)
\end{aligned}
\label{eq:act_aply}
\end{equation}

\paragraph{Transition function.} In this study, state transitions $\mathcal{T}: \mathcal{S} \times \mathcal{A} \rightarrow P(s_{t+1}|s_t, a_t)$ are integrated in \textbf{RideGym} and are not explicitly modeled in our method.

\paragraph{Reward function.} After calculating the coupon assignment $j^*$ for each passenger's request by Eq.~\ref{eq:relaxedLagrangian-adjj} with $\lambda_t$ at step $t$, the reward function $r \in \mathcal{R}: \mathcal{S} \times \mathcal{A} \rightarrow \mathbb{R}$ is defined in Eq.~\ref{eq:reward}, which reflects the order completions resulting from this coupon assignment.
\begin{equation}
r_t(s_t,a_t) = \sum_{i=0}^{N_t-1}z^t_{ij^*(\lambda_t(s_t,a_t))}
\label{eq:reward}
\end{equation}
\paragraph{Penalty function.} At each time step $t$, issuing coupons increases the probability of order completions, while also consuming the budget. We thus define a penalty function $p \in \mathcal{P}: \mathcal{S} \times \mathcal{A} \rightarrow \mathbb{R}$, demonstrated in Eq.~\ref{eq:penalty}, to record the budget consumption.
\begin{equation}
p_t(s_t,a_t) = \sum_{i=0}^{N_t-1}z^t_{ij^*(\lambda_t(s_t,a_t))}g^t_i d_{j^*(\lambda_t(s_t,a_t))}
\label{eq:penalty}
\end{equation}
\paragraph{Full-Achievement function.}
In our scenario, the constraint is a predetermined budget rate over a specified period $[t_{start}, t_{end}]$, rather than a fixed budget amount. Compliance with this constraint depends not only on the expenditure, but also on the total GMV achieved by our coupon allocation strategy. Therefore, we employ Eq.~\ref{eq:CR} to integrate past cumulative subsidy expenditures, GMV gains (from $t_{start}$ to $t$), and forecasts up to $t_{end}$ to assess adherence to the budget rate constraint. We propose a Full-Achievement function, demonstrated in Eq.~\ref{eq:f_t}, to balance the total achievement of order completions and budget rate compliance. Note that the total order completions for the entire period are normalized by the previously solved optimal order completions using the ternary search method in $[t_{start}, t_{end}]$, denoted by $R^*$.
\begin{equation}
CR_t = \frac{\sum_{s=0}^{t-1}p^{obs}_s+ \sum_{s=t}^{T-1}p_s(s_s,a_s)}{\sum_{s=0}^{t-1}r^{obs}_s + \sum_{s=t}^{T-1}r_s(s_s,a_s)}
\label{eq:CR}
\end{equation}
\begin{equation}
Constraint\_Penalty = e^{max(\frac{CR_t}{CR^*}-1,\;0)}-1
\label{eq:cp}
\end{equation}
\begin{equation}
F_t = \frac{\sum_{s=0}^{t-1}r^{obs}_s + \sum_{s=t}^{T-1}r_s(s_t,a_t)}{R^*} - Constraint\_Penalty
\label{eq:f_t}
\end{equation}

\subsubsection{Parameters Initialization of Decision Function}
First, we pretrain the fundamental models and initialize $\lambda$ on the dataset $\mathcal{D}_{t_{start}-L}^{t_{start}-1}$. The fundamental models, $\mathcal{W}$, $\mathcal{F}^{(in)}$ and $\mathcal{F}^{(out)}$, are used to estimate the parameters $w_{ij}$, $f_{ij}^{(in)}$, and $f_{ij}^{(out)}$, respectively, in Eq.~\ref{eq:relaxedLagrangian-adjj} during decision making. This allows us to solve for $\lambda^*$ on $\mathcal{D}_{t_{start}-L}^{t_{start}-1}$ via ternary search to maximize Eq.~\ref{eq:relaxedLagrangian}. The resulting $\lambda^*$ is then used by the actor to adjust the system at the initial time step. 

\subsubsection{Actor \& Critic}
In this study, we adopt the Actor-Critic framework~\cite{haarnoja2018soft}. At each time slot $t$, the actor $\pi_{\theta}$, parameterized by $\theta$, is a neural network that takes the current state $s_t$ as input and outputs the mean $\mu_t$ and standard deviation $\sigma_t$ of a Gaussian distribution. An action $a_t$ is then sampled from this distribution. The action $a_t$ adjusts the previous $\lambda_{t-1}$ by Eq.~\ref{eq:act_aply}. To stabilize $\lambda$ updates, we adjust $\lambda$ using Eq.\ref{eq:lambda_adj}, where $\delta \lambda_{t-1}$ denotes the previous change margin and $\lambda^o_t$ represents the $\lambda$ originally modified by $a_t$; this adjusted value is then utilized in Eq.\ref{eq:relaxedLagrangian-adjj} to determine the optimal coupon allocation for each request in the current time slot. The critic $Q_{\gamma}(\cdot, \cdot)$ estimates the Full-Achievement $F_t$ based on the state-action pair $(s_t, a_t)$. 
\begin{equation}
\lambda_t = \lambda_{t-1} + (\lambda^o_t - \lambda_{t-1}) / (1 + \delta\lambda_{t-1})
\label{eq:lambda_adj}
\end{equation} 

\subsection{Fast Competition Adaption (FCA)}
At each time step $t$, to promptly respond to fluctuations in (\textit{IRR}) induced by sporadic price adjustments by competitors, we propose modeling the initial \textit{IRR} landscape for each candidate coupon as a set of Beta distributions. This facilitates capturing dynamic changes through Bayesian posterior updates. Under the Homogeneous Competitors Assumption (Assumption~\ref{asmt:id}), Proposition~\ref{prop: beta} holds, validating the use of Beta distributions to model the initial \textit{IRR} distribution.
\newtheorem{assumption}{Assumption}
\begin{assumption}
    (Homogeneous Competitor Assumption). \textit{All RSPs belong to the same service category (e.g., economy), which implies that when they join the RHA at $t=0$, the prices $p_m(\mathbf{x})$ offered by all RSPs for requests with identical characteristics $\mathbf{x}$ follow an independent and identically distributed continuous distribution with a cumulative distribution function (CDF) $F_X$.}
\label{asmt:id} 
\end{assumption}

\begin{proposition}
Let $M$ denote the total number of RSPs for a given request $i$. The CDF of the $K(\textbf{x}_i)$-th lowest price, denoted as $F_{X_{(K(\textbf{x}_i))}}$ follows a Beta distribution (proof provided in Appendix D). It follows directly that the \textit{IRR} for request $i$ at $t=0$ is governed by:
\begin{equation}
    Pr(p_m(\mathbf{x}_i) \leq X_{(K(\textbf{x}_i))}) \sim Beta(K(\textbf{x}_i), M+1-K(\textbf{x}_i)).
\end{equation}
\label{prop: beta}
\end{proposition}
In the following, we demonstrate how the conjugate properties of the Beta distribution facilitate rapid Bayesian posterior updates at time step $t$, enabling swift adaptation to competitors' price adjustments at the previous time step.

\paragraph{Clustering.}
A key challenge is that samples are not identical across time steps, so the observed In-Range outcomes at time $t-1$ cannot be directly applied to individual samples at time $t$. To address this, we employ the K-Means algorithm~\cite{hartigan1979algorithm} to cluster requests with similar features, establishing a mapping $f: \mathbb{X} \to \mathbb{C} = \{0,\ldots,S-1\} \subseteq \mathbb{N}$ from the feature space to the cluster labels. This allows us to implement a Bayesian posterior update at time $t$ based on observations from the corresponding cluster at $t-1$. We denote the most recent parameters of the Beta distribution for cluster $c$ as $\alpha_{c,d}$ and $\beta_{c,d}$, for each $d\in \mathbf{d}$.

\paragraph{Initialization of \textit{IRR} Prior.}
In our framework, the tracking of the \textit{IRR} distribution commences at time step $t_{start}$. We average the predictions from model $\mathcal{W}$ over the dataset $\mathcal{D}_{t_{start}-L}^{t_{start}-1}$, stratified by the previously defined clustering labels and coupon levels. This yields the initial parameters $\alpha_{c,d}^{t_{start}}$ and $\beta_{c,d}^{t_{start}}$ for each cluster $c$ and coupon $d$.  We store these in a dictionary $\mathcal{I}$, with the cluster centers as keys and the corresponding $(\alpha_{c,d}^{t_{start}}, \beta_{c,d}^{t_{start}})$ as values to track the \textit{IRR} landscape. 

\paragraph{Bayesian Posterior Update.}
Assuming intra-cluster homogeneity during the FCA process (with an acceptable loss of precision), we model the number of In-Range samples in each cluster under each candidate coupon as a binomial. Specifically, let $N^{t-1}_{c,d}$ denote the number of samples in cluster $c$ at time $t-1$ that received coupon $d$, and $N^{in,t-1}_{c,d}$ the number of those samples that are In-Range. Given the Binomial–Beta conjugacy, the posterior at time $t$ also follows to a Beta distribution (proof provided in Appendix E). As a result, the parameters at $t$ can be updated by
\begin{equation}
    \begin{aligned}
    \alpha^{t}_{c,d} &= N_{c,d}^{in,t-1}+\alpha^{t-1}_{c,d}
    \\ \beta^{t}_{c,d} &= N_{c,d}^{t-1} -N_{c,d}^{in,t-1}+\beta^{t-1}_{c,d}.\\
    \end{aligned}
    \label{eq:cluster_update}
\end{equation}
 To address potential stochastic noise in Beta distribution updates due to limited samples per time step, we introduce a window size parameter, $l$, aggregating In-Range observations from the preceding $l$ time steps (Setting $l=1$ is consistent with Eq.~\ref{eq:cluster_update}). 

\subsection{Integration of FCA in RLA}
This section describes the integration of FCA into the RLA actor's decision-making process. FCA influences two key aspects of decision-making:
\paragraph{Augmentation of $s_t$.}
 At time step $t$, the general \textit{IRR} status tracked by FCA under coupon $d$, represented by $\frac{1}{S} \sum_{c} \alpha_{c,d}^{t}$ and $\frac{1}{S} \sum_{c} \beta_{c,d}^{t}$, is incorporated into the state vector $s_t$. This allows the actor to adjust $\lambda_{t-1}$ to $\lambda_{t}$ based on the updated \textit{IRR} information.
\paragraph{Refinement of $w$.}
To preserve heterogeneity across samples, the original predictions, denoted as $\alpha^{t,ori}_{(\mathcal{W}(\textbf{x},d))}$ and $\beta^{t,ori}_{(\mathcal{W}(\textbf{x},d))}$, from $\mathcal{W}$ for each individual sample are retained but adjusted using the latest parameters tracked by FCA. The refined predictions for a sample in cluster $c$ at time step $t$ are as follows:
\begin{equation}
    \begin{aligned}
    \alpha^{t}_{\textbf{x},d} &= \alpha^{t,ori}_{(\mathcal{W}(\textbf{x},d))} + \alpha^{t}_{c=f(x),d}
    \\ \beta^{t}_{\textbf{x},d} &= \beta^{t,ori}_{(\mathcal{W}(\textbf{x},d))} + \beta^{t}_{c=f(x),d}
    \end{aligned}
    \label{eq:update}
\end{equation}
These updated estimates, along with the new $\lambda_t$, are incorporated into Eq.~\ref{eq:relaxedLagrangian-adjj} to determine the coupon allocation at time step $t$.

After the coupons are issued, \textbf{RideGym} returns the actual In-Range and order completion results for that time step. FCA collects the $N^{in,t}_{c,d}$ and $N^{t}_{c,d}$ to update $\alpha_{c,d}$ and $\beta_{c,d}$ for each cluster and observed coupon. Any unexpected price adjustments by competitors are captured by the updated values of $\alpha_{c,d}$ and $\beta_{c,d}$. The complete process of our FCA-RL is demonstrated in Algo.~\ref{algr:fca-rl}.


\begin{algorithm}[hbt!]
\DontPrintSemicolon
  \SetAlgoLined
  \KwIn{$B$; $\mathcal{W}$; $\mathcal{F}^{(in)}$; $\mathcal{F}^{(out)}$; $\lambda^*_0$; $R^*$; $CR^*$; $\mathcal{I}$; RideGym $\mathcal{E}$; $\mathbf{d}$; Batch size $b$;} 
  \KwOut{$\pi_{\theta}$; $Q_{\gamma}$; $\mathcal{I}$}
  \While{not convergent}{
        Set $\mathbf{\lambda}_{0} = \mathbf{\lambda}^*_{0} + \mathcal{\epsilon}$ ;\;
        Observe state $\mathbf{s}_0$ from $\mathcal{E}$;\;
        Set $R = 0$; $C = 0$;\;
        \For{$t = 0$ to $T$}{
            Get an action $\mathbf{a}_t \sim \pi_{\theta}(\mathbf{s}_t)$;\;
            Get $\lambda_t$ by Eq.~\ref{eq:act_aply};\;
            Get $\mathbf{\alpha}^{t,ori} \in \mathcal{R}^{N_t\times|\textbf{d}|}$ and Get $\mathbf{\beta}^{t,ori}\in \mathcal{R}^{N_t\times|\textbf{d}|}$ estimated by $\mathcal{W}$;\;
            Get $\mathbf{f}^{(in)}\in \mathcal{R}^{N_t\times|\textbf{d}|}$, $\mathbf{f}^{(out)} \in \mathcal{R}^{N_t\times|\textbf{d}|}$ estimated by $\mathcal{F}^{(in)}$ and $\mathcal{F}^{(out)}$;\;
            Update $\mathbf{\alpha}^{t}\in \mathcal{R}^{N_t\times|\textbf{d}|}$ and $\mathbf{\beta}^{t} \in \mathcal{R}^{N_t\times|\textbf{d}|}$ by Eq.~\ref{eq:update};\;
            Sample $\mathbf{w}^t \in \mathcal{R}^{N_t\times|\textbf{d}|}$ from $\textbf{Beta}(\mathbf{\alpha}^{t},\mathbf{\beta}^{t})$;\;
            Calculate coupon allocation by Eq.~\ref{eq:relaxedLagrangian-adjj};\;
            Get reward $r_t$, $p_t$, $s_{t+1}$, $\{N^t_{c,d}\}^{|\mathbb{C}|\times|\mathbf{d}|}$ and $\{N^{in,t}_{c,d}\}^{|\mathbb{C}|\times|\mathbf{d}|}$ from $\mathcal{E}$;\;
            Augment $s_{t+1}$ by $\{\frac{1}{S} \sum_{c} \alpha_{c,d}^{t}\}^{|\textbf{d}|}$ and $\{\frac{1}{S} \sum_{c} \beta_{c,d}^{t}\}^{|\textbf{d}|}$;\;
            Update dictionary $\mathcal{I}$ by Eq.~\ref{eq:cluster_update} ;\;
            Set $R = R + r_t$; $C = C + p_t$; Set $V=0$; Set $Z=0$;\;
            \For{$k=t+1$ to $T$}{
            Get $r_k$, $p_k$ by repeat execution of line $6$ to $14$;\;
            Set $V=V+r_k$; Set $Z = Z+p_k$;\;
            }
            Set $CR_t = \frac{C+Z}{R+V}$; Set $CP = e^{max(\frac{CR_t}{CR^*}-1,0)}-1$ ;\;
            Set $F_t = \frac{(R+V)}{R^*} - CP$;\;
            Store $(s_t, a_t, F_t)$ in $\mathbf{\mathcal{M}}$ ;\;    
        }
        
    Sample $b$ of tuples from $\mathbf{\mathcal{M}}$;\;
    Update Critic by minimizing the loss $L_{Critic} = \frac{1}{b}\sum_i(F^i - Q(s^i, a^i))^2$\;
    Update Actor by maximizing the PPO loss $L^{\text{CLIP}}(\theta) = \hat{\mathbb{E}}_t \left[ \min \left( r_t(\theta) \hat{F}_t, \text{clip}(r_t(\theta), 1 - \epsilon, 1 + \epsilon) \hat{F}_t \right) \right| r_t(\theta) = \frac{\pi_\theta(a_t | s_t)}{\pi_{\theta_{\text{old}}}(a_t | s_t)}]$\;
  }
  \caption{FCA-RL Algorithm}
  \label{algr:fca-rl}
\end{algorithm}

\section{RideGym}
\label{sec:simulation-system}

\begin{figure}[t!]
    \centering
    \includegraphics[width=0.5\linewidth]{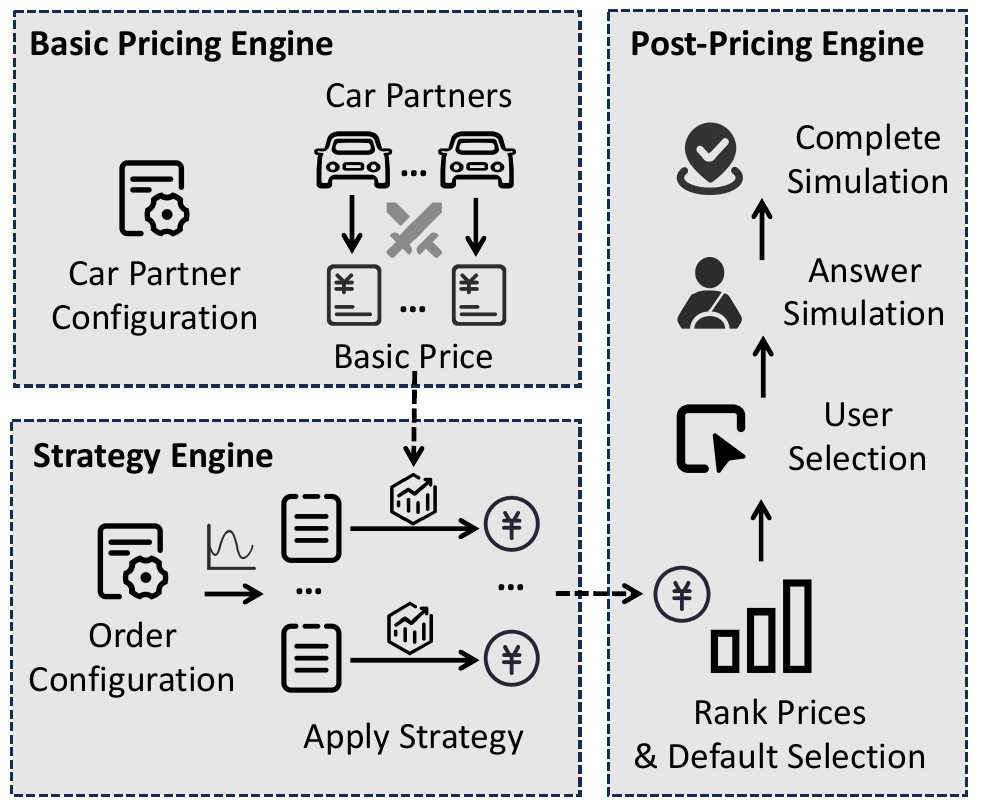}
    \caption{The pipeline of RideGym. 
    }
    \label{fig:gym-pipeline}
\end{figure}

Implementing new subsidy strategies directly in online environments carries the risk of financial loss. To mitigate this risk, we propose the \textbf{RideGym} simulation system, which models the complete operations of a ride-hailing aggregator. As depicted in Figure~\ref{fig:gym-pipeline}, RideGym consists of three core components: the Basic Pricing Engine, the Strategy Engine, and the Post-Pricing Engine. This simulation framework enables the evaluation of different subsidy strategies under diverse market conditions without the potential losses associated with real-world implementation.

\paragraph{Basic Pricing Engine.} 
In this engine, the base price of each order is determined by a configurable price per mile. Price adjustments by one RSP affect others within the RHA. To simulate market competition, we introduce a stochastic price adjustment mechanism by configuring the frequency, lower and upper bounds of each competitor's price adjustment. The actual adjustment amount is sampled from a uniform distribution defined by these bounds. The adjusted price for the m-th RSP is calculated as: $p_{m}^{\text{aft}} = p_{m} \times (1 + a)$, where $a$ is the sampled adjustment amount.

\paragraph{Strategy Engine.} The Strategy Engine generates orders for each time slot, maintaining the desired order count distribution by using multiple normal distributions to approximate the actual distribution. Each RSP can then apply its configurable coupon strategy to assign coupons to the generated orders. 


\paragraph{Post-Pricing Engine.} 
The post-pricing engine processes the quoted prices from all RSPs, ranks them in ascending order, and simulates both automatic and passenger selection. Automatic selection by RHA employs a top-$K$ mechanism, while the number of RSPs selected by a passenger, denoted by $K'$, is calculated as follows: given a list of quoted prices $(p_1, p_2, ..., p_M)$ for $M$ RSPs, 
$$K' = \text{Clip}\left(K + \log_{b}\left(\text{density}(p_1, ..., p_M) + b^{-K}\right), 1, M\right)$$

where $K$ is the number of RSPs selected by default, $b$ is an adjustable base, and $\text{density}(\cdot)$ calculates the density of the price list. The clipping function ensures that $K'$ stays within the range $[1, M]$. After the order is sent by passengers with selected RSPs, each order is assigned a random supply factor $s \in [0, 1]$ to model supply conditions. Assuming service capabilities $(tp_1, tp_2, ..., tp_M)$ for the $M$ RSPs, the probability of the $i$-th RSP answering an order is: $P(\text{answer}=i \mid s) = s \cdot \frac{tp_i}{\sum_{j=1}^M tp_j}$.
The no answer rate of an order is: $P(\text{no answer} \mid s) = 1 - \sum_{j=1}^M P(\text{answer}=j \mid s)$. Finally, the simulation includes a random normal distribution to model potential order cancellations. 



\section{Evaluation}
\label{sec:evaluation}

\subsection{Evaluation Setup}

\paragraph{Dataset Detail.}
Our RideGym platform generates four different scenarios (referred to as Scenes 1 to 4) under varying configurations. Each scenario consists of three datasets: Pre-train, Train, and Test, with the Pre-train and Train datasets being consistent across all scenarios. In Scenes 1 to 3,  price adjustments by other RSPs through changes in base prices occur with increasing frequency. These adjustments directly influence an RSP's ranking within the RHA form and affect the IRR. To assess the robustness of our method in stable pricing environments, we introduce Scene-4, where no RSPs change their prices. See Appendix B for more details on the configuration of the different scenes.

\paragraph{Training and Evaluation Details.}
In this section, we first describe the training details for each scene. For the FCA-RL model, both $\mathcal{W}$ and $f^{(in)}$ are MLPs trained on the Pre-train dataset. In our implementation, we omit the prediction of $f^{(out)}_{ij}$, assuming that RSPs placed Out-Range have a negligible probability of being selected by passengers. These pre-trained models are then used to infer the Test dataset without further weight optimization. Further details can be found in Appendix A.1.

\paragraph{Baseline Details.}
We introduce three baselines to analyze the coupon strategy in RHA:
(i) \textbf{Optimal Method(OPT)}: This approach utilizes the ground truth of the completion label to derive the optimal solution. (ii) \textbf{Primal-Dual Method-A(PDM-A)}: A traditional technique that implements uplift modeling the predictions of the completion probability $z_{ij}$ in an end-to-end fashion. It then solves a constrained optimization problem using the Primal-Dual Method. (iii) \textbf{PDM-S}:This method explicitly models $w_{ij}$ and $f^{(in)}_{ij}$, similar to our FCA-RL approach. However, it employs the Primal-Dual Method to derive the solutions without real-time adjustments to the IRR and $\lambda$. To construct proper estimates for the parameters in their optimal decision function, we train MLPs to predict $w_{ij}$, $f^{(in)}_{ij}$ for PDM-S, and $z_{ij}$ for PDM-A. All baselines are solved on the same Train dataset as FCA-RL. Their solutions remain unchanged during evaluation on the Test set. 

\paragraph{Evaluation Metric.}
Here, we introduce three metrics for analyzing algorithms.

\textbf{Cost Rate Error, CRE} quantifies the precision of a coupon strategy in controlling costs. It is defined as the difference between the actual coupon cost rate and the target budget rate, denoted as $CR^*$. Mathematically, the total CRE is expressed as:
\begin{equation}
    CRE = \Big \vert\frac{\sum_{i}\text{cost}_i}{\sum_{i}\text{gmv}_i} - CR^*\Big \vert
\end{equation}

\textbf{Finish Return Of Interest, FROI} assesses the effect of coupon strategies on increasing order volumes, where $F$, $F_0$ are the count of finished orders with/without strategy, $C$ is the subsidy cost of the subsidy strategy, $A$ and $A_0$ are the Average Selling Price (ASP) with/without strategy.
\begin{equation}
    FROI = \frac{F-F_0}{\frac{A_0}{A}C}
\end{equation}

\textbf{Reinforcement Learning Reward, RLR} gauges the order increasing effect and the control ability at the same time:
\begin{equation}
    RLR = \frac{\sum_{i=1}^T r_i^{obs}}{R^*}{} - (e^{max(\frac{CR}{CR^*}-1,0)}-1)
\end{equation}\textbf{}


\subsection{Empirical Results}
Here, we summarize our evaluation target to the following research questions. Experiments run on a machine with a 16-core Intel CPU, 40GB RAM, and GeForce RTX 2080 Ti GPU. Results are averaged over five random seeds. 

\paragraph{RQ1: How does FCA-RL compare to the other baselines in terms of performance?}



We present three key metrics for Scene-$1$ to $3$ across all baselines. Since the ground truth for In-Range and order completion results is generated by our RideGym, we obtain the optimal $\lambda^*$ using ternary search. As shown in Table~\ref{table:str-pref-3scene-transposed}, our RL-FCA significantly outperforms all baselines on these metrics. It achieves smaller budget rate errors, indicated by the CRE column (up and down arrows correspond to over- and underspending), and higher money efficiency in scenarios with different price adjustment frequencies. In competitive pricing scenarios such as Scene-2 and Scene-3, RL-FCA limits budget rate errors to within $0.3$ percentage points (pp), surpassing the second-best method, PDM-S, by reductions of $0.6$ pp and $0.4$ pp. While occasionally overspending compared to PDM-S, RL-FCA demonstrates higher efficiency, with FROI increases of $0.1$\% in Scene-1 and $3.6$\% in Scene-3, defying the typical economic principle that marginal benefits decrease as costs increase.

\begin{table}[t]
\centering
\caption{Strategy Performance on Three Scenes}
\label{table:str-pref-3scene-transposed}
\begin{tabular}{l *{3}{ccc}}
\toprule
 & \multicolumn{3}{c}{Scene-1} & \multicolumn{3}{c}{Scene-2} & \multicolumn{3}{c}{Scene-3} \\
\cmidrule(lr){2-4} \cmidrule(lr){5-7} \cmidrule(lr){8-10}
\textbf{Method} & \textbf{CRE} & \textbf{FROI} & \textbf{RLR} & \textbf{CRE} & \textbf{FROI} & \textbf{RLR} & \textbf{CRE} & \textbf{FROI} & \textbf{RLR} \\
\midrule
Optimal & $0.000$ & $1.663$ & $0.48$ & $0.001$ & $1.388$ & $0.283$ & $0.000$ & $1.366$ & $0.476$ \\
PDM-A   & $(\uparrow)0.023$ & $1.305$ & $-0.111$ & $(\uparrow)0.032$ & $1.016$ & $-0.499$ & $(\uparrow)0.020$ & $1.062$ & $-0.149$ \\
PDM-S   & $(\downarrow)0.002$ & $1.575$ & $0.127$ & $(\uparrow)0.012$ & $1.228$ & $0.040$ & $(\downarrow)0.007$ & $1.262$ & $0.104$ \\
RL-FCA  & $(\downarrow)\mathbf{0.001}$ & $\mathbf{1.578}$ & $\mathbf{0.168}$ & $(\downarrow)\mathbf{0.006}$ & $\mathbf{1.454}$ & $\mathbf{0.167}$ & $(\downarrow)\mathbf{0.003}$ & $\mathbf{1.308}$ & $\mathbf{0.208}$ \\
\bottomrule
\end{tabular}
\end{table}






\paragraph{RQ2: What is the effect of the FCA component?}



We conducted ablation experiments on FCA modules across various scenarios (Scene-$1$ to Scene-$4$, shown in Table~\ref{table:ablation-fca}). Additionally, we analyzed the impact of window size on FCA effectiveness in the most competitive pricing scenario (Scene-$3$), with results presented in Table~\ref{table:window-size-fca}. Table~\ref{table:ablation-fca} shows that the FCA module reduces budget rate execution errors in all scenes. It outperforms standard RL without \textit{IRR} adaptation in competitive scenarios (Scene-$2$ and $3$) by $32.2$\% and $77.4$\% in RLR. However, in stable price scenarios (Scene-$1$ and $4$), FCA offers no advantage, as the stable distribution may hinder RL training convergence. As shown in Table~\ref{table:window-size-fca}, testing different window lengths in Scene-$3$ revealed that longer windows improve RL training, with minimal differences beyond a length of $20$. We therefore chose a final window length of $24$.



\begin{table}[t]
\parbox[t]{.42\linewidth}{
\centering
\caption{Ablation Study for FCA}
\begin{tabularx}{0.5\textwidth}{@{}ccXXX@{}}
\toprule
&\textbf{FCA} 
&\textbf{CRE} 
&\textbf{FROI}
&\textbf{RLR}\\
\midrule
\multirow{2.5}{*}{Scene-1} & 
w
& $(\downarrow)\mathbf{0.001}$ & $1.578$ & $0.168$ \\
\cmidrule(lr){2-5}
& 
w/o
& $(\downarrow)0.003$ & $\mathbf{1.653}$ & $\mathbf{0.197}$ \\
\midrule
\multirow{2.5}{*}{Scene-2} & w  & $(\downarrow)\mathbf{0.001}$ & $1.316$ & $\mathbf{0.119}$ \\
\cmidrule(lr){2-5}
& w/o  &  $(\downarrow)0.011$ & $\mathbf{1.500}$ &  $0.090$\\
\midrule
\multirow{2.5}{*}{Scene-3} & w  & $(\downarrow)\mathbf{0.003}$ & $1.308$ & $\mathbf{0.208}$\\
\cmidrule(lr){2-5}
& w/o  & $(\downarrow)0.02$ & $\mathbf{1.466}$ & $0.117$  \\
\midrule
\multirow{2.5}{*}{Scene-4} & w  & $(\downarrow)\mathbf{0.002}$ & $1.362$ & $0.125$ \\
\cmidrule(lr){2-5}
& w/o  & $(\downarrow)0.008$ & $\mathbf{1.453}$ & $\mathbf{0.195}$ \\
\bottomrule
\end{tabularx}
\label{table:ablation-fca}
}
\hspace{4em}
\parbox[t]{.42\linewidth}{
\centering
\caption{Study for the Window Size} 
\begin{tabularx}{0.45\textwidth}{@{}>{\centering\arraybackslash}X X X X@{}}
\toprule
\textbf{WS} & \textbf{CRE} & \textbf{FROI} & \textbf{RLR}   \\ 
\midrule
$1$           & $(\downarrow)0.017$ & $1.398$ & $0.133$ \\ 
\midrule
$3$           & $(\downarrow)0.019$ & $1.397$ & $0.143$ \\ 
\midrule
$10$          & $(\downarrow)0.007$ & $1.346$ & $0.182$ \\ 
\midrule
$15$          & $(\downarrow)0.005$ & $1.325$ & $0.192$ \\ 
\midrule
$20$          & $(\downarrow)0.003$ & $1.308$ & $0.203$ \\ 
\midrule
$24$          & $(\downarrow)\mathbf{0.003}$ & $1.308$ & $\mathbf{0.208}$ \\ 
\midrule
$30$          & $(\downarrow)\mathbf{0.003}$ & $1.308$ & $\mathbf{0.208}$ \\
\bottomrule
\end{tabularx}
\label{table:window-size-fca}
}
\end{table}

\paragraph{RQ3: How does our RL-FCA outperform the others?}

\begin{figure}[t]
	\centering
\begin{tabularx}{\textwidth}{@{}c@{\hspace{1.5mm}}c@{\hspace{1.5mm}}c@{}} 
\includegraphics[width=0.323\textwidth,height=0.2\textheight,keepaspectratio]{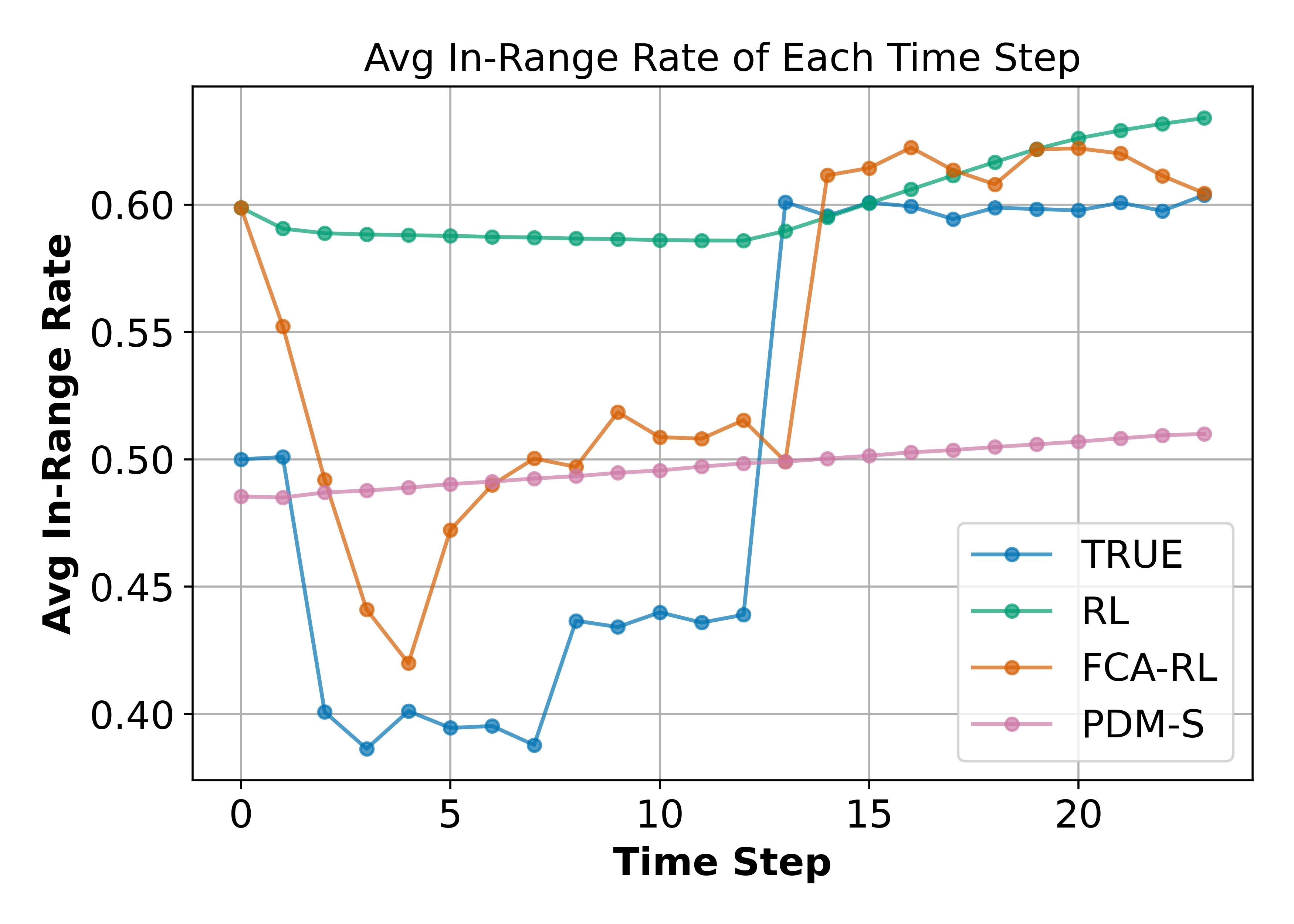} &
\includegraphics[width=0.323\textwidth,height=0.2\textheight,keepaspectratio]{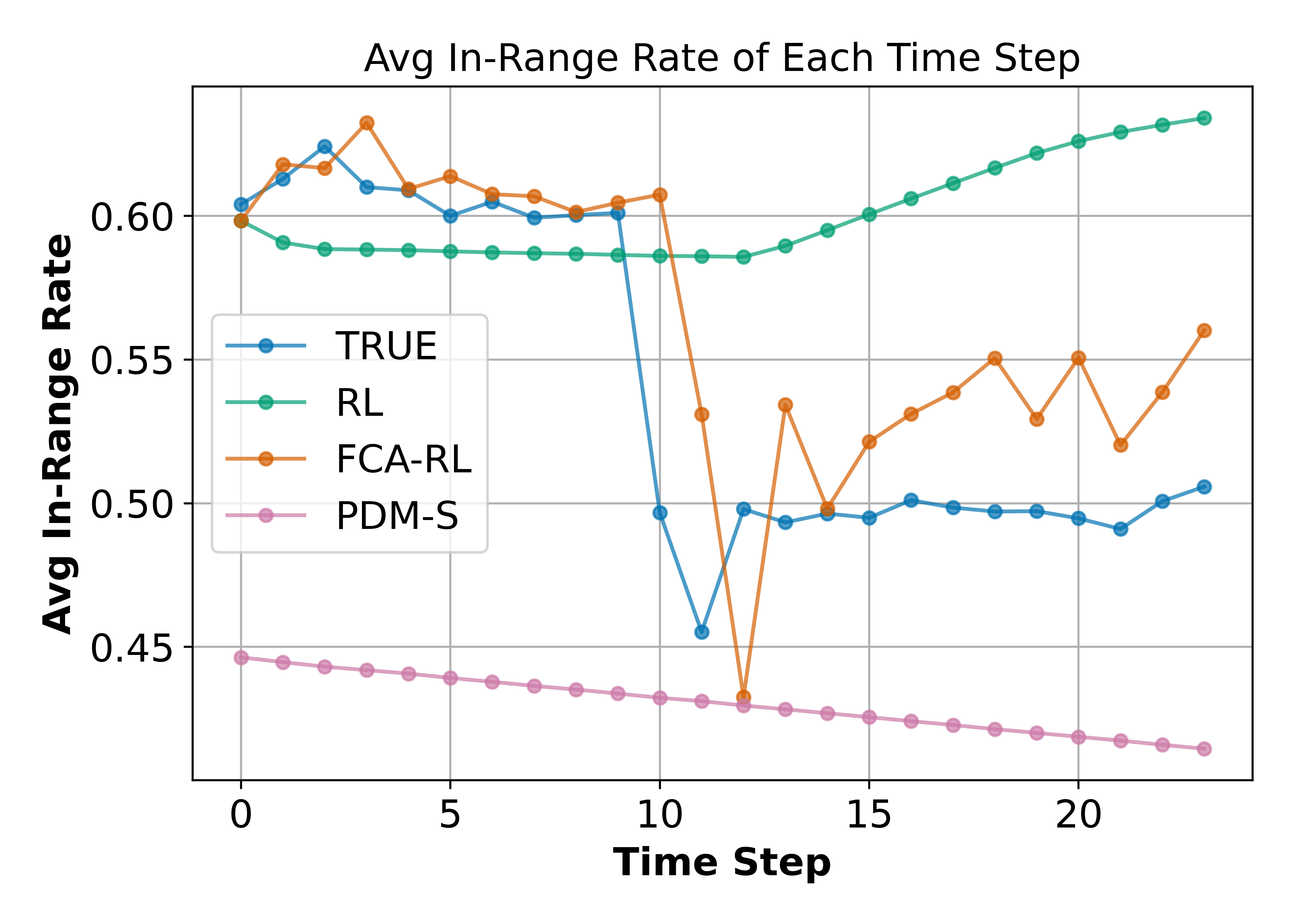} &
\includegraphics[width=0.323\textwidth,height=0.2\textheight,keepaspectratio]{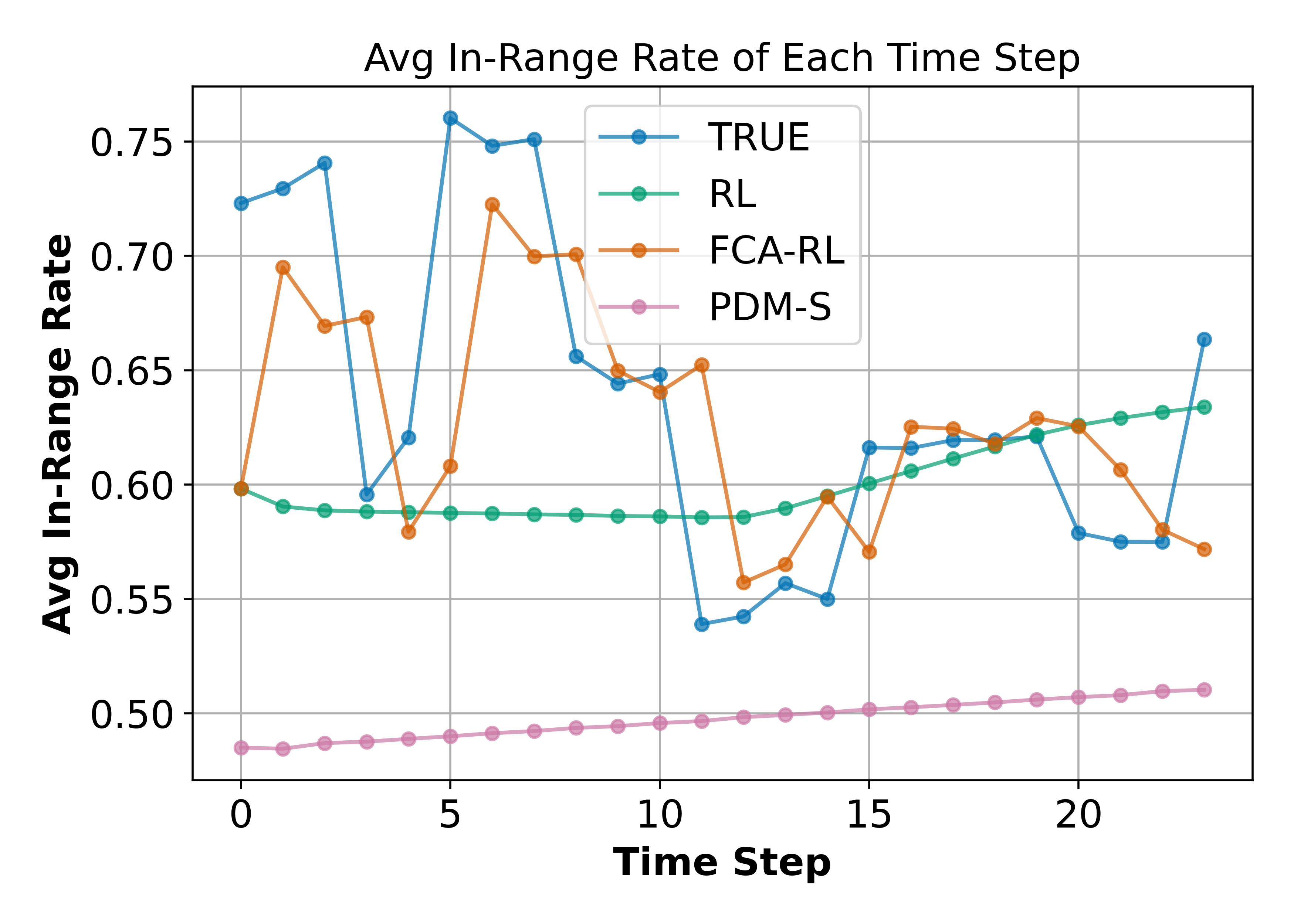} \\
scene-1 & scene-2 & scene-3 
    \end{tabularx}
	\caption{Average \textit{IRR} adjustment for each method over each time step}
	\label{fig:wr_dist}
\end{figure}
\begin{figure}[t]
	\centering
\begin{tabularx}{\textwidth}{@{}c@{\hspace{1.5mm}}c@{\hspace{1.5mm}}c@{}} 
\includegraphics[width=0.323\textwidth,height=0.2\textheight,keepaspectratio]{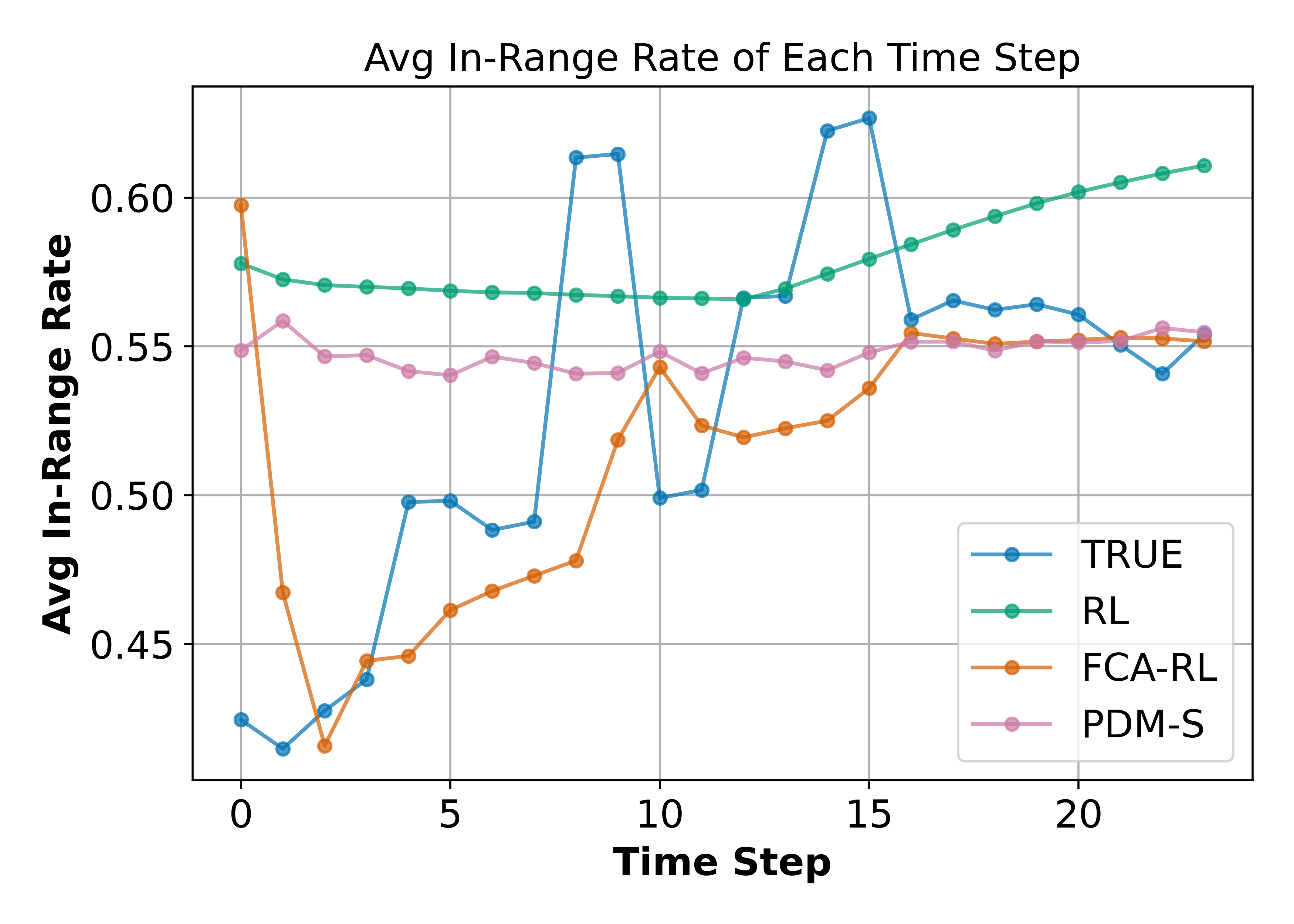} &
\includegraphics[width=0.323\textwidth,height=0.2\textheight,keepaspectratio]{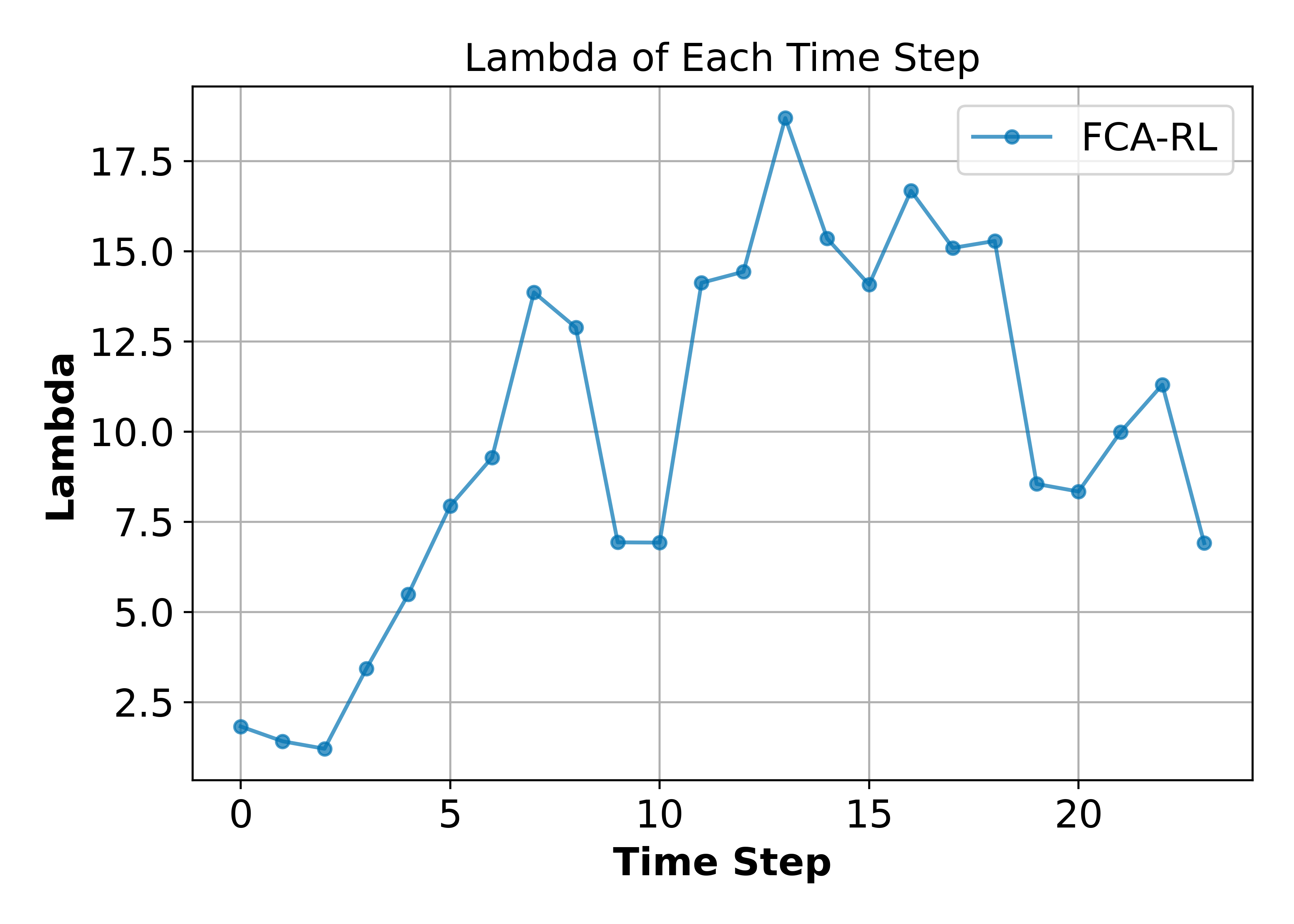} &
\includegraphics[width=0.323\textwidth,height=0.2\textheight,keepaspectratio]{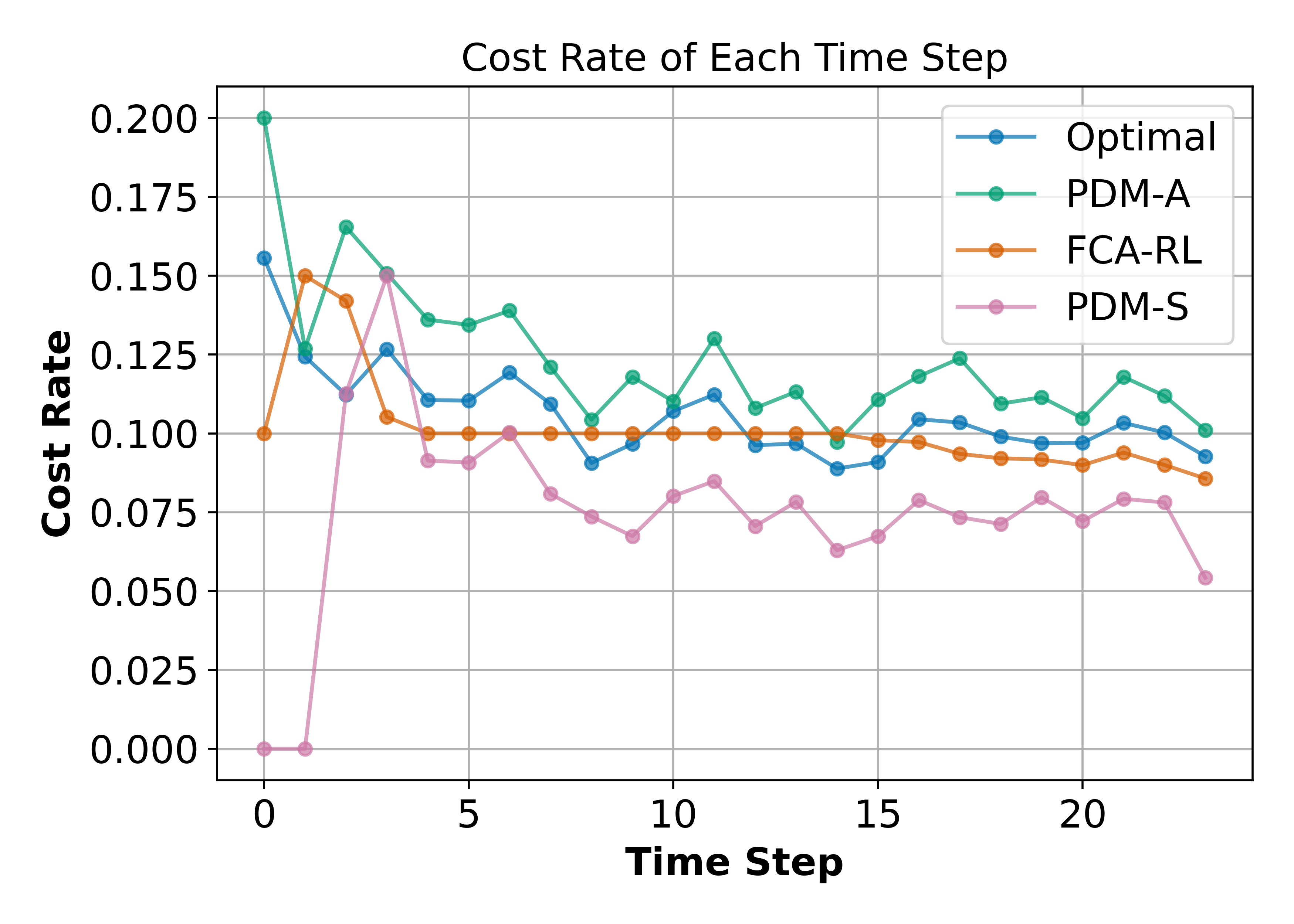} \\
Average \textit{IRR} & $\lambda$ adjusted by RL agent & Cost Rate 
    \end{tabularx}
	\caption{Visualization of the performance of our FCA-RL method in Scene-3}
	\label{fig:s3}
\end{figure}

Our RL-FCA adapts to market changes by adjusting the Lagrange multiplier ($\lambda$) in real-time. Figure~\ref{fig:wr_dist} illustrates how the true \textit{IRR} distribution shifts due to price adjustments and compares our method's quick adaptation to others relying on training set distribution. To highlight FCA's response to \textit{IRR} distribution changes, we show results with a window size of 1, where FCA is more sensitive but also introduces greater fluctuations in RL training. Figure~\ref{fig:s3} illustrates how the RL agent adjusts the $\lambda$ parameter at each time step in response to changes in the \textit{IRR} distribution (first two columns). The third column displays the budget rate implementations for each method. FCA-RL promptly adjusts $\lambda$, achieving a smoother budget rate and a spending profile that closely aligns with the optimal strategy compared to other methods.

\section{Conclusion}
We propose FCA-RL, a reinforcement learning-based approach to optimize order acquisition for RSPs under price competition in RHA. Its effectiveness is demonstrated through experiments on RideGym, our self-developed simulation system for managing the full lifecycle of ride requests. In this study, we primarily address temporary data distribution shifts caused by price competition. However, the potential passenger response to coupons and long-term supply-demand dynamics remain underexplored, which we aim to address in future work.
\label{sec:conclusion}

\bibliographystyle{splncs04}
\bibliography{95article_reference}

%
%
%
%

\end{document}